%% file: main.tex
\documentclass[letterpaper,10pt,conference]{ieeeconf}

\IEEEoverridecommandlockouts
\input{preamble}

\usepackage[hidelinks]{hyperref}
\usepackage[capitalize]{cleveref}

\title{\LARGE \bf
LHNav: Long-Horizon Humanoid Navigation via Multirate Coordination and Physics-Grounded Action Data
}

\author{Xueyang Wang$^{1,2}$, Zehong Shen$^{2,*}$, Dingkun Wei$^{1,2}$, Huaijin Pi$^{2,3}$,\\
Changpeng Wang$^{1,2}$, Xing Zhu$^{2}$, Hujun Bao$^{1}$, Yujun Shen$^{2,\dagger}$, Xiaowei Zhou$^{1,\dagger}$\\[0.4em]
\small $^{*}$Project Lead \quad $^{\dagger}$Corresponding Authors\\[0.3em]
\small $^{1}$Zhejiang University \quad $^{2}$Robbyant \quad $^{3}$The University of Hong Kong%
\thanks{Xueyang Wang, Dingkun Wei, Huaijin Pi, and Changpeng Wang conducted this work during their internships at Ant Group. This work was supported by Ant Group Research Intern Program.}}

\hypersetup{
  pdftitle={LHNav: Long-Horizon Humanoid Navigation via Multirate Coordination and Physics-Grounded Action Data},
  pdfauthor={Xueyang Wang, Zehong Shen, Dingkun Wei, Huaijin Pi, Changpeng Wang, Xing Zhu, Hujun Bao, Yujun Shen, Xiaowei Zhou}
}

\begin{document}

\IEEEaftertitletext{%
  \begin{center}
    \includegraphics[width=\textwidth]{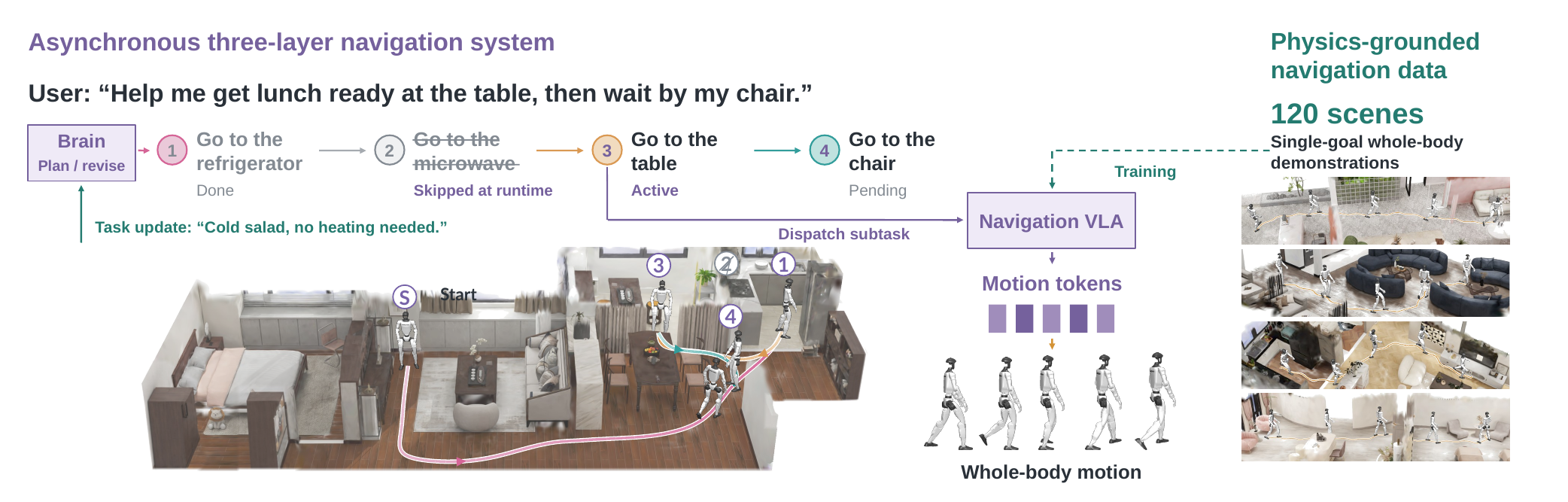}\endgraf
    \captionof{figure}{\textbf{\method combines an asynchronous three-layer
    navigation system with physics-grounded navigation data.}
    The Brain maintains and revises a navigation plan during execution,
    dispatching single-goal instructions to a VLA that predicts whole-body
    motion tokens. The scene shows the initial navigation sequence, while
    the task row illustrates removing an unnecessary pending visit after
    a task update. The data engine covers 120 scenes, with 100 for training
    and 20 for held-out testing. Single-goal training demonstrations pair
    visual observations with motion tokens.}
    \label{fig:teaser}
  \end{center}
}

\maketitle
\thispagestyle{empty}
\pagestyle{empty}

\input{sections/0_abstract}
\input{sections/1_introduction}
\input{sections/2_related_work}
\input{sections/3_method}
\input{sections/4_data_engine}
\input{sections/4_experiments}
\input{sections/5_conclusion}

{\small
\bibliographystyle{IEEEtran}
\bibliography{main}
}


\end{document}

%% file: preamble.tex
\usepackage{graphicx}        
\usepackage{capt-of}         
\usepackage{amsmath,amssymb} 
\usepackage{booktabs}        
\usepackage{multirow}        
\usepackage{tabularx}        
\usepackage{adjustbox}       
\usepackage{placeins}        
\usepackage{xcolor}          
\usepackage{xspace}          
\usepackage{cite}            
\usepackage{url}             
\usepackage{microtype}

\makeatletter
\let\NAT@parse\undefined
\makeatother

\makeatletter
\let\nhnav@makecaption\@makecaption
\long\def\@makecaption#1#2{%
  \ifx\@captype\@IEEEtablestring
    \vskip\abovecaptionskip
    {\normalfont\normalsize\noindent #1: #2\par}%
  \else
    \nhnav@makecaption{#1}{#2}%
  \fi}
\makeatother

\newcommand{\method}{\textsc{LHNav}\xspace}



%% file: sections/0_abstract.tex
\begin{abstract}

We study humanoid navigation across multiple destinations under long-horizon language instructions. 
This task is challenging because robots must maintain progress across successive goals while continuously producing natural, stable whole-body motion through confined spaces. 
However, task decomposition in existing methods does not by itself ensure progress throughout execution, while waypoint-based navigation does not guarantee natural, route-faithful whole-body motion. 
Our key idea is to organize navigation into three complementary roles operating at different time scales: a \emph{Brain} for task reasoning, a \emph{Cerebellum} for local motion generation, and a \emph{Brainstem} for whole-body execution.
We realize this idea in \method, an asynchronous system that uses progress and arrival feedback to coordinate navigation goals, VLA motion-token prediction, and whole-body control. 
To train the Cerebellum for natural, route-faithful motion, we develop a physics-grounded data engine that improves physical tracking of motion references along collision-aware routes and pairs the resulting demonstrations with rendered visual observations.
Together, these designs allow task planning to proceed alongside continuous movement and extend single-goal navigation to successive destinations with natural, route-faithful whole-body motion.
In simulation, our navigation VLA outperforms OmniNav and DualVLN in single-goal success and motion smoothness, with fewer collisions and falls.
Using a VLA trained only on single-goal demonstrations, \method achieves a 75.56\% success rate on long-horizon navigation tasks. 
Deployment on the Unitree G1 further demonstrates single-goal navigation and zero-shot long-horizon task execution in real-world environments, including layouts with repositioned objects.
Project website: https://robbyant-research.github.io/LHNav/.
\end{abstract}

%% file: sections/1_introduction.tex
\section{Introduction}
\label{sec:introduction}

Humanoid robots assisting people in everyday environments need to follow long-horizon language instructions that involve multiple destinations and task dependencies~\cite{ahn2022saycan,zhou2026holoagent0}.
For example, preparing a meal may require visiting the refrigerator, chopping board, stove, and dining table in an appropriate order.
In this work, we focus on the navigation required by such instructions, where a humanoid must infer navigation goals and maintain progress across destinations without step-by-step guidance.
This setting requires both task-level coordination across successive goals and natural, stable whole-body motion along long routes through confined spaces.

Long-horizon humanoid navigation is challenging because task progress must remain consistent with continuous execution across multiple destinations.
Existing methods couple navigation decisions to reasoning~\cite{cai2025clcotnav,gao2025octonav,zuo2026fantasyvln}, while asynchronous and hierarchical approaches allow task planning and action to proceed at different time scales~\cite{wei2025dualvln,huang2026ticvla,su2026sagenav,zhou2026holoagent0}.
However, separating these processes does not itself ensure coherent task progression: goal transitions must reflect actual execution progress while navigation continues.
A second challenge is reconciling route constraints with natural, stable whole-body motion, particularly in confined spaces.
Waypoints, trajectories, and velocity commands describe intended movement but leave its whole-body realization to downstream controllers~\cite{krantz2021waypoint,huang2026ticvla,zhou2026holoagent0}, so a collision-free route does not directly guarantee collision-free physical execution.
Recent methods introduce richer body-level actions and multi-stage behaviors~\cite{ding2025humanoidvla,wang2026vlk,rayyan2026fetchman,kim2026legs}, yet maintaining natural motion while limiting position and heading errors over long, confined routes remains challenging.

We present \method, combining asynchronous navigation with a physics-grounded data engine for long-horizon humanoid navigation.
Inspired by the human brain, it comprises three layers: a \emph{Brain} that maintains task context and dispatches goals, a \emph{Cerebellum} that predicts motion tokens with auxiliary waypoint-based geometric supervision, and a \emph{Brainstem} that executes whole-body motion through a frozen SONIC decoder~\cite{luo2025sonic}.
Progress and arrival feedback coordinate successive goals, allowing a single-goal policy to execute navigation while task planning proceeds asynchronously.
Our data engine uses ARDY~\cite{zhao2026ardy} to generate natural motion references along collision-aware routes and trains the SONIC encoder with additional reference-root and velocity information for accurate tracking through the frozen decoder.
Rendering closed-loop rollouts in reconfigurable Gaussian--mesh scenes~\cite{kerbl20233dgs} pairs visual observations with synchronized motion-token targets for VLA training.
Together, asynchronous coordination and physically executed supervision support natural, route-faithful whole-body navigation across multiple goals.
Joint training on these simulated navigation demonstrations and real teleoperated manipulation data further enables navigation followed by tabletop pushing on Unitree G1.

Our contributions are threefold:
\begin{enumerate}
    \item An asynchronous multirate system integrating task reasoning, motion-token prediction, and whole-body execution, using progress-aware transitions that extend a single-goal policy to long-horizon humanoid navigation.
    \item A scalable, physics-grounded data engine that improves global motion-reference tracking through a privileged encoder and frozen SONIC decoder, generating execution-aligned navigation demonstrations in Gaussian--mesh scenes.
    \item Experimental validation demonstrating higher navigation success in unseen simulation scenes, smoother motion, fewer collisions, and faster long-horizon completion, alongside real-world single-goal navigation and zero-shot long-horizon navigation.
\end{enumerate}

%% file: sections/2_related_work.tex
\section{Related Work}
\label{sec:related}

\textbf{From navigation commands to whole-body motion.}
Navigation policies typically output motion primitives~\cite{krantz2020vlnce}, metric commands~\cite{zhang2024navid,cheng2024navila}, or local waypoints and trajectories~\cite{krantz2021waypoint,sridhar2023nomad,cai2025navdp}.
While natural for wheeled robots, these interfaces rely on downstream controllers for actuation and, on humanoids, whole-body coordination.
Richer interfaces expose body targets: LEGS combines wrist poses with base commands~\cite{kim2026legs}, FetchMan uses base height and upper-body joint targets~\cite{rayyan2026fetchman}, and VLK predicts root-and-joint trajectories with contacts~\cite{wang2026vlk}.
Others encode motion as pose tokens~\cite{ding2025humanoidvla}, learned latents~\cite{xue2025leverb}, or FSQ tokens~\cite{luo2025sonic}.
SONIC decodes such tokens with proprioceptive feedback and supports direct VLA prediction.
We adopt this interface for navigation, motivating supervision aligned with executed humanoid motion.

\begin{figure}[t]
    \addtolength{\abovecaptionskip}{-5pt}
    \centering
    \includegraphics[width=1.0\columnwidth]{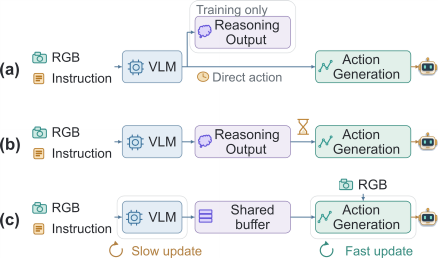}
    \caption{Representative reasoning--action coupling patterns:
    (a) reasoning supervision; (b) think then act;
    (c) asynchronous reasoning and control.}
    \label{fig:reasoning-comparison}
\end{figure}

\input{figures/method_figure}

\textbf{Physics-grounded navigation supervision.}
Navigation data draw on instruction--route datasets~\cite{anderson2018vln}, simulators~\cite{savva2019habitat}, Gaussian scene representations~\cite{kerbl20233dgs,interiorgs2025}, and geometric experts that synthesize routes~\cite{cai2025navdp,abotn02026}.
These provide visual context and geometric routes, but humanoid supervision must also specify how the body executes them.
VLK synthesizes whole-body kinematics in reconstructed scenes~\cite{wang2026vlk}, while LEGS and FetchMan collect control commands from physical rollouts~\cite{kim2026legs,rayyan2026fetchman}.
Tracking reveals failures and route deviations missed by kinematic references; LEGS therefore retains only successful episodes~\cite{kim2026legs}.
Our data engine targets global route fidelity and pairs rendered executed states with the post-FSQ tokens consumed during rollout, matching the deployment action interface.

\textbf{Reasoning and execution for long-horizon tasks.}
Long-horizon navigation couples semantic planning with local execution.
\cref{fig:reasoning-comparison} summarizes representative strategies.
FantasyVLN uses reasoning supervision only during training~\cite{zuo2026fantasyvln}, whereas NavGPT and CL-CoTNav reason online before acting~\cite{zhou2023navgpt,cai2025clcotnav}.
OctoNav and AwareVLN reduce overhead by adapting reasoning frequency or switching between reasoning and action modes~\cite{gao2025octonav,guo2026awarevln}.
OmniNav connects planning and execution through subgoals~\cite{xue2025omninav}, while DualVLN enables asynchronous action prediction from cached goals and current observations~\cite{wei2025dualvln}; TIC-VLA further models latency and egomotion~\cite{huang2026ticvla}.
Task coordination also requires execution feedback: SAGE-Nav updates semantic waypoints upon completion or deadlock~\cite{su2026sagenav}, while Uni-LaViRA, Being-0, and HoloAgent-0 organize multistage execution through subgoal or skill interfaces~\cite{ding2026unilavira,yuan2025being0,zhou2026holoagent0}.
We use asynchronous execution, with short instructions and target images linking task planning to single-goal navigation.
Arrival verification and progress feedback guide advancement and recovery; motion tokens drive high-rate proprioceptive whole-body control.

%% file: figures/method_figure.tex
\begin{figure*}[!t]
    \vspace*{5pt}
    \addtolength{\abovecaptionskip}{-5pt}
    \centering
    \includegraphics[width=\textwidth]{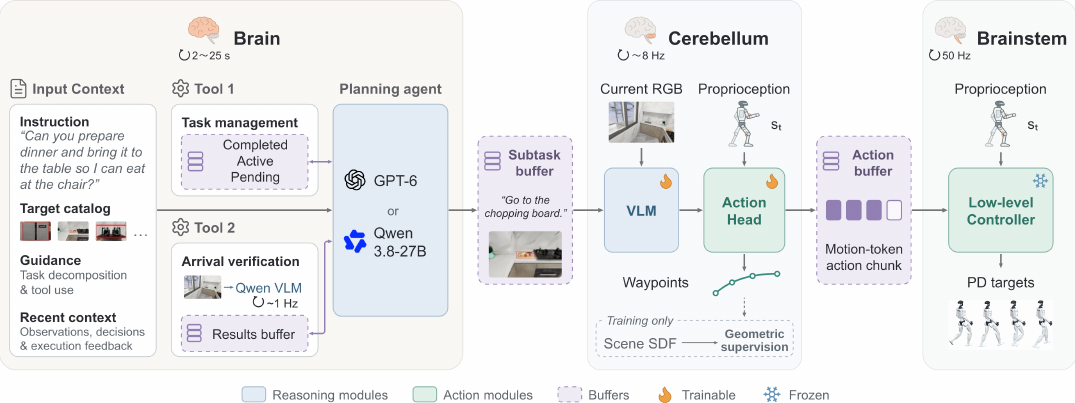}
    \caption{\textbf{Our asynchronous three-layer navigation system.}
    The Brain uses task management (Tool~1) to maintain an explicit plan and
    revise pending goals during execution, with goal-completion feedback from
    arrival verification (Tool~2).
    A subtask buffer supplies an instruction and target image to the
    Cerebellum, which uses current RGB and proprioceptive state $s_t$ to jointly
    predict motion-token chunks and auxiliary waypoints at approximately 8~Hz.
    An action buffer supplies the latest motion-token chunk to the Brainstem's
    frozen SONIC controller, which combines it with proprioception to produce
    joint targets at 50~Hz for PD control.
    The independent arrival-verification VLM updates the results buffer at
    up to 1~Hz.
    Geometric supervision encourages waypoint traversability during training.}
    \label{fig:method}
\end{figure*}

%% file: sections/3_method.tex
\section{Method}
\label{sec:method}

We consider a G1 humanoid robot operating in a scene with a catalog of
reference images depicting candidate destinations. Given a long-horizon
language instruction, \method selects the required destinations from this
catalog and navigates the robot to them in sequence.
As shown in \cref{fig:method}, the Brain decomposes the instruction into
navigation subtasks (\cref{sec:method:architecture}), the Cerebellum predicts local motion tokens, and the
Brainstem decodes them into whole-body actions (\cref{sec:method:vla}). All three layers execute asynchronously.

\subsection{Brain: Agent with Tools}
\label{sec:method:architecture}

The Brain coordinates long-horizon navigation through a planning agent with task-management and arrival-verification tools, tracking progress, revising the remaining plan, and deciding when to advance.

The planning agent receives the full instruction and target catalog, together with task-decomposition and tool-use guidance and recent observations, decisions, and execution feedback.
It can be implemented with GPT-6 or Qwen3.8-27B.
The task management tool (Tool~1 in \cref{fig:method}) maintains an explicit plan of completed, active, and pending navigation subtasks.
The agent uses this tool to create the plan and revise pending subtasks during execution using new observations, task updates, and execution feedback.
Revisions can remove goals that are no longer needed, while preserving completed visits and the active goal.
The arrival verification tool provides goal-completion feedback through an independent Qwen VLM, which checks arrival at up to 1~Hz using the active subtask, target reference image, and temporal observations.
Diagnostic analyses and subtask-specific arrival judgments are written to a shared buffer for the planning agent to access.
At a navigation stop, the runtime harness blocks new VLA requests and discards outstanding predictions.
It records completion after verified arrival, motion-plan exhaustion, and odometry-confirmed root stability, then dispatches the next subtask from the updated plan following the external stage-completion event.
At each stage, the Brain supplies the Cerebellum VLA with a simple textual instruction specifying the active navigation goal, such as ``go to the table.''

\input{figures/data_engine_figure}

\subsection{Cerebellum and Brainstem: Humanoid VLN with Geometric Supervision}
\label{sec:method:vla}

Given a navigation subtask from the Brain, the Cerebellum uses a humanoid
VLA to predict whole-body motion tokens from visual observations and
proprioception. The Brainstem decodes these tokens into joint targets for
execution.

\textbf{Joint action and waypoint prediction.}
We build on the fine-tuning framework for GR00T N1.7 and
SONIC~\cite{nvidia2026sonicvla}, retaining its vision--language backbone and
flow-matching action model. Inspired by successful VLN methods that use
waypoint-based action representations~\cite{krantz2021waypoint,xue2025omninav,wei2025dualvln},
we introduce geometric supervision
into motion-token learning by jointly predicting motion tokens and auxiliary
waypoints with a shared DiT.
Given a subtask instruction $\ell_t$, target image $I_g$, current
RGB observation $I_t$, and proprioceptive state $s_t$, the policy predicts
\begin{equation}
 (\widehat Z_t,\widehat W_t)
 =\pi_\theta(\ell_t,I_g,I_t,s_t),
 \label{eq:vla-joint-prediction}
\end{equation}
where $\widehat Z_t\in\mathbb R^{H\times64}$ contains motion tokens and
$\widehat W_t\in\mathbb R^{H\times5}$ contains waypoints, with $H=40$.
Each waypoint encodes a planar position in the observation-time robot frame,
heading through its sine and cosine, and an arrival indicator. Motion tokens
and waypoints are indexed by time and path distance, respectively, and are
not aligned frame by frame.

\textbf{Geometric supervision.}
To improve waypoint traversability, we introduce SDF-based geometric supervision.
For simplicity, we restrict the signed distance field $d_{\mathrm{sdf}}$ to
the $xy$ plane and precompute it from the background
traversable mesh (BgMesh) and the current configuration of foreground movable
obstacle meshes (FgMesh). The field is positive within the resulting
traversable region and negative outside.
We apply this supervision to waypoint predictions obtained after the full
denoising process. Their planar positions are transformed into scene
coordinates for bilinear SDF queries. The minimum signed clearance over valid
waypoints determines the penalty:
\begin{equation}
 \begin{aligned}
 c_{\min}&=\min_{i\in\mathcal V}
 d_{\mathrm{sdf}}\!\left(T_t\widehat w_i^{xy}\right),\\
 \mathcal L_{\mathrm{sdf}}&=\eta\log\!\left(1+\exp(-c_{\min}/\eta)\right),
 \end{aligned}
 \label{eq:vla-navmesh}
\end{equation}
where $T_t$ maps local positions to scene coordinates, $\mathcal V$ indexes
valid waypoints, and $\eta$ controls the transition scale. We use minimum
clearance to prevent localized boundary violations from being diluted by
averaging over safe waypoints. Softplus smoothly approximates this violation
penalty.

\textbf{Training and deployment.}
The training objective combines joint flow matching with geometric supervision:
\begin{equation}
 \mathcal L=\mathcal L_{\mathrm{joint\text{-}flow}}(Z,W)
 +\lambda_{\mathrm{sdf}}\mathcal L_{\mathrm{sdf}}.
 \label{eq:vla-objective}
\end{equation}
The Cerebellum updates motion-token chunks at approximately 8~Hz using 4
denoising steps, while the frozen SONIC decoder~\cite{luo2025sonic} executes
the current chunk at 50~Hz. Returned chunks skip the prefix corresponding to
steps executed during inference and replace the unexecuted tail without blending.

%% file: figures/data_engine_figure.tex
\begin{figure*}[t]
    \vspace*{5pt}
    \addtolength{\abovecaptionskip}{-5pt}
    \centering
    \includegraphics[width=\textwidth]{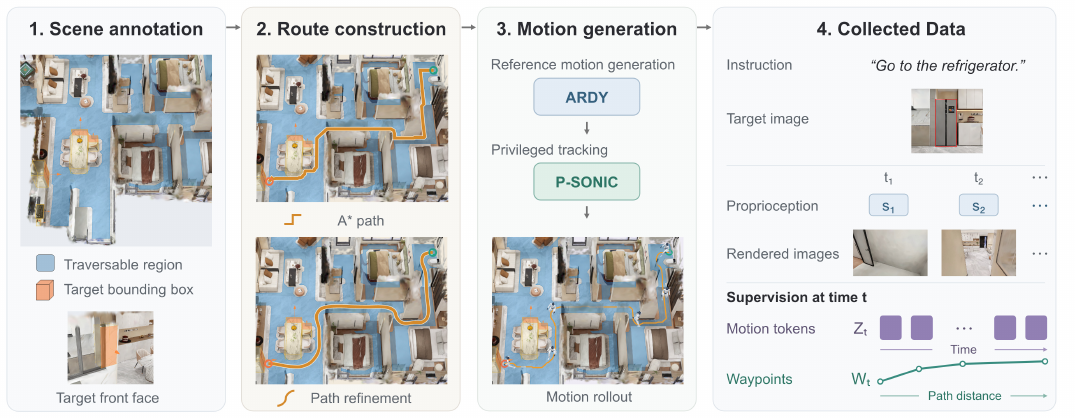}
    \caption{\textbf{Example of navigation data generation for
    ``Go to the refrigerator.''}}
    \label{fig:data-engine}
\end{figure*}

%% file: sections/4_data_engine.tex
\section{Humanoid Navigation Data Engine}
\label{sec:data-engine}

We design a physics-grounded data engine for training and evaluating
humanoid navigation policies.
As shown in \cref{fig:data-engine}, it converts target-conditioned routes into
physically executed demonstrations that provide visual observations,
motion tokens, and waypoint labels.

\subsection{Scene and Route Construction}
\label{sec:data-engine:planning}
Generating visual navigation demonstrations requires environments that support realistic rendering.
For training, we use 100 photorealistic scenes from InteriorGS~\cite{interiorgs2025} as the scene backgrounds.
To plan routes within these environments, we manually annotate a background mesh (BgMesh) that identifies where the robot can move.
To vary target and obstacle layouts, we augment the Gaussian backgrounds with movable foreground meshes (FgMesh), forming GS+Mesh scenes.
Route planning then uses the traversable region determined by BgMesh and the current foreground configuration.

Target-conditioned navigation also requires specifying where the robot should stop and how it should face the target.
We annotate target bounding boxes and facing directions, then select an accessible approach pose outside each target's bounding box, prioritizing its front side and orienting the robot toward it.
To generate routes from different initial conditions, we randomly sample a start position and heading in the same connected traversable region and connect the start to the approach pose using 2D A*.
To maintain obstacle clearance and avoid abrupt changes in direction, we refine each route for clearance, smoothness, and length.
This produces a target-conditioned navigation path connecting the sampled start position to the approach position, together with the initial and terminal headings.

\subsection{Route-Conditioned Motion Generation and Tracking}
\label{sec:data-engine:motion}

Given a route, we use the ARDY motion generator~\cite{zhao2026ardy} to
synthesize a whole-body reference motion at 50~Hz.
Throughout generation, the prompt is ``A person is walking.'' Soft path and
heading constraints and target standing poses at both endpoints guide route
following and motion transitions.

Physical tracking errors can cause executed demonstrations to deviate from
their planned routes. We therefore improve global reference tracking to
obtain route-faithful demonstrations for navigation-policy training.
SONIC~\cite{luo2025sonic} encodes future reference motion into quantized
motion latents and decodes them with robot proprioception into joint actions.
As shown in \cref{fig:psonic-tracking}, the original SONIC can accumulate
global drift during execution, causing the executed trajectory to deviate
from the ARDY reference and leaving a visible position offset at the endpoint. Our
privileged SONIC (P-SONIC) augments the encoder with future reference pelvis
position differences and recent body-frame linear velocities:
\begin{equation}
 p_t=\operatorname{concat}\!\left(
 \{R_t^\top(x^{\mathrm{ref}}_{t+\tau_j}-x_t)\}_{j=1}^{10},
 \{v^{\mathrm{body}}_{t-j}\}_{j=9}^{0}\right).
 \label{eq:psonic-privilege}
\end{equation}
Here, $x_t$ and $R_t$ are the current robot root position and rotation,
$x^{\mathrm{ref}}$ is the reference pelvis position, $\tau_j$ specifies a
future sampling offset, and $v^{\mathrm{body}}$ is the actual body-frame
linear velocity. We fine-tune the
encoder with PPO in IsaacLab for 40K training iterations across 262K parallel
environments, using global tracking and relative body-pose objectives while
freezing the FSQ quantizer and decoder.

For offline data generation, P-SONIC tracks the references in MuJoCo. We
retain rollouts whose root tilt-angle error remains below $25^\circ$ as the
no-fall criterion, recording executed robot states and the post-quantization
motion tokens supplied to the decoder. Keeping the decoder frozen preserves
the decoding interface at deployment. The VLA learns to predict the motion
latents labeled by P-SONIC, and its predictions are executed by the same
decoder without privileged inputs.

\input{figures/psonic_tracking_figure}

Across 184,990 shared routes in 100 scenes, our pipeline reduces the
proportion of executed path length outside the traversable region from
3.318\% to 1.841\% and joint-position jerk by 44.7\%, compared with guiding
SONIC along the same routes using velocity commands toward a look-ahead
point. This yields smoother navigation demonstrations with fewer
traversable-region violations.

\subsection{Execution-Aligned Rendering and Supervision}
\label{sec:data-engine:rendering}

We replay recorded robot states to render first-person observations using
the robot-mounted camera configuration. For GS+Mesh scenes,
the Gaussian background and FgMesh are rendered from the same camera and
composited using depth. Each episode shares an instruction and target image
across its observation and proprioception sequences. At each observation,
supervision includes a temporal motion-token chunk and waypoints sampled by
path distance from the executed trajectory, expressed in the current robot
frame.

%% file: figures/psonic_tracking_figure.tex
\begin{figure}[!t]
    \centering
    \includegraphics[width=\columnwidth]{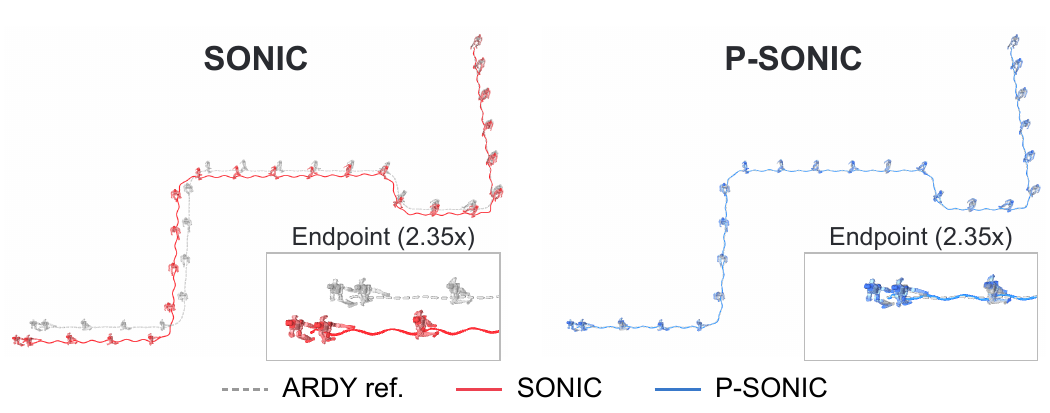}
    \caption{\textbf{Reference tracking on the route in
    \cref{fig:data-engine}.} Motion runs from upper right to lower left;
    insets enlarge the endpoints.}
    \label{fig:psonic-tracking}
\end{figure}

%% file: sections/4_experiments.tex
\section{Experiments}
\label{sec:experiments}

\begingroup
\raggedbottom
\setlength{\intextsep}{6pt}
\setlength{\textfloatsep}{8pt}

We evaluate the Navigation Cerebellum on single-goal navigation and the
complete system on long-horizon tasks in simulation, followed by real-robot
tests in household and reconfigured environments and a qualitative
navigation--interaction example. Ablations examine privileged motion tracking,
action representations, geometric supervision, and VLA update frequency.

\subsection{Experimental Setup}
\label{sec:exp:principle}

\paragraph{Datasets}
We select 100 training scenes and 20 held-out test scenes from
InteriorGS~\cite{interiorgs2025}. Our data engine produces 185,401 single-goal
training demonstrations, and we evaluate all methods on 522 episodes with
identical starts and goals in the held-out scenes. For long-horizon
experiments, we use 20 scenes from the training split, with 14,410 single-goal
demonstrations for policy training and 45 evaluation tasks requiring at least
three ordered goal visits.

\paragraph{Training Settings}
For single-goal evaluation, we fine-tune OmniNav~\cite{xue2025omninav} and
DualVLN~\cite{wei2025dualvln} from their released checkpoints, and our
Navigation Cerebellum from pretrained GR00T N1.7
weights~\cite{nvidia2026sonicvla}, on the same 100-scene dataset.
For long-horizon evaluation, we train a separate Cerebellum from the same
initialization using single-goal demonstrations from the 20-scene subset.
Zero-shot denotes execution of long-horizon instructions unseen during
navigation-policy training on single-goal demonstrations.
All navigation policies are trained for 100K steps with a global batch size
of 64 on eight NVIDIA H20 GPUs.

\paragraph{Evaluation Settings}
All methods use SONIC~\cite{luo2025sonic} for whole-body control.
Motion-token execution does not require odometry-based trajectory tracking.
Under the same execution setting, waypoint-based policies issue velocity
commands along the initial direction of each predicted trajectory, refreshed
with every new prediction. The commanded speed is 0.6~m/s, matching the
walking speed used by ARDY during data collection.

\paragraph{Hardware Setup}
We deploy on a Unitree G1 with a forward-facing Orbbec Gemini335L camera
secured to the head using a custom 3D-printed mount
(\cref{fig:hardware-setup}). The camera provides egocentric RGB observations
for task reasoning, arrival verification, and local navigation.

\begin{figure}[t]
    \vspace*{5pt}
    \addtolength{\abovecaptionskip}{-5pt}
    \centering
    \includegraphics[width=\columnwidth]{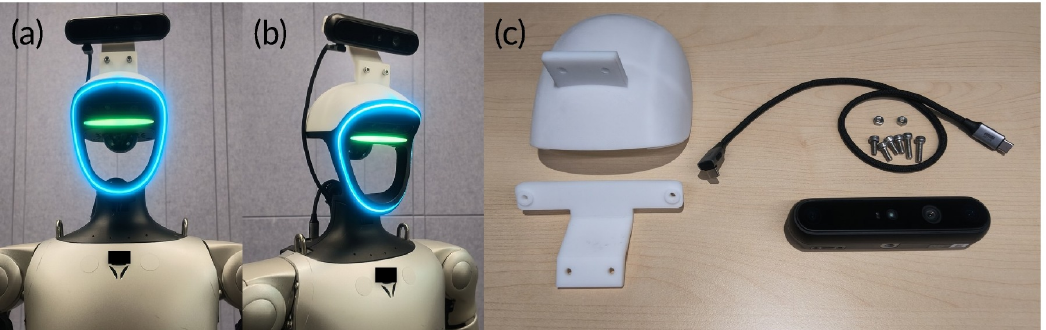}
    \caption{\textbf{Real-robot platform and camera mounting.}
    (a,b) Front and oblique views of Unitree G1 with a forward-facing
    Orbbec Gemini335L camera mounted above the head for egocentric RGB observations.
    (c) Camera assembly components, including the custom 3D-printed mount,
    fastening hardware, and connection cable.}
    \label{fig:hardware-setup}
\end{figure}

\paragraph{Metrics}
SR$_{0.5}$/SR$_{1.5}$ measures success below 0.5/1.5~m final goal distance
without falls. Falls counts non-collision tracking falls. NE measures final planar
distance to the goal point (last target box for long-horizon tasks).
CR averages the root-path fraction outside traversable space. Joint-position
jerk ($J_p$) averages unfiltered third-order finite differences of 29 world-frame
joint positions sampled at 50~Hz over joints, time, and episodes.
Long-horizon Task SR$_{1.5}$ requires matching the
prescribed target sequence, with every recorded completion position within
1.5~m horizontally of its corresponding target box. Progress is the fraction completed before the
first error; OSR counts ordered entry into all target regions regardless of
stopping. Time averages successful-task duration, including planning,
navigation, handoffs, and dwell.

\subsection{Single-Goal Navigation in Simulation}
\label{sec:exp:m1}

Under these execution settings, our pipeline achieves higher success, lower
CR and joint-position jerk, and fewer falls than adapted waypoint baselines
in unseen scenes (\cref{tab:m1-short-sim}), supporting accurate navigation
with smooth, stable motion. \Cref{fig:sim-single-goal} shows two example rollouts.

\input{tables/m1_short_sim_results}

\begin{figure*}[t]
    \vspace*{5pt}
    \addtolength{\abovecaptionskip}{-5pt}
    \centering
    {\normalfont\fontsize{8}{9.5}\selectfont\color{black}
    \makebox[\textwidth][l]{\makebox[0.5\textwidth][l]{(a) Television}(b) Pet Nest}\par\nobreak
    \includegraphics[width=\textwidth]{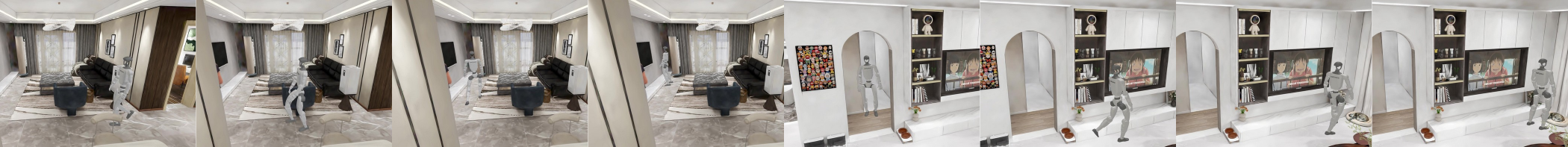}\par}
    \caption{\textbf{Single-goal navigation to a television and a pet nest.}
    Each half shows four chronological frames from a selected successful
    rollout, including intermediate walking and final goal positioning.}
    \label{fig:sim-single-goal}
\end{figure*}

\subsection{Long-Horizon Navigation in Simulation}
\label{sec:exp:system}

All variants share the single-goal VLA and SONIC decoder, with Qwen3.8-27B
for VLM-based variants and GPT-6-Astra (low reasoning effort) for Codex.
The full system achieves 75.56\% Task SR$_{1.5}$ (\cref{tab:system}; example
in \cref{fig:sim-long-horizon}). Removing task management, including the plan,
completed-visit history, and textual and visual execution memory, reduces
success to 0\%; removing arrival verification and its retry and turning
recovery reduces it to 55.6\%. Codex success improves from 55.6\% to 64.4\%
with Qwen's task-decomposition guidance through the harness, and to 71.1\%
with both tools. Asynchronous planning and verification overlap navigation
rather than run between attempts, achieving comparable success to the
synchronous variant with lower mean completion time over each variant's
successful tasks.

\input{tables/system_results}

\begin{figure*}[t]
    \vspace*{5pt}
    \addtolength{\abovecaptionskip}{-5pt}
    \centering
    \includegraphics[width=\textwidth]{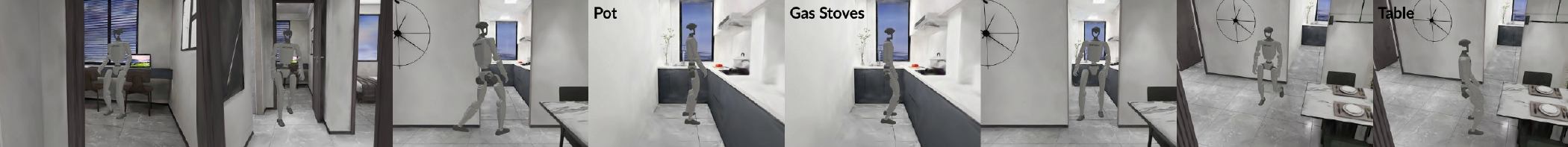}
    \caption{\textbf{Sequential goal reaching in a simulated long-horizon task.}
    Instruction: ``I'm craving something hot for dinner. Please cook a simple
    meal in the pot on the gas stoves and serve it on the table.''
    Frames run chronologically from left to right, retaining intermediate
    walking and labeling the three goal locations.
    The rollout executes navigation and arrival holds, not cooking or serving.}
    \label{fig:sim-long-horizon}
\end{figure*}

\subsection{Real-Robot Navigation}
\label{sec:exp:m3}

\begin{figure*}[t]
    \centering
    {\normalfont\fontsize{8}{9.5}\selectfont\color{black}
    \makebox[\textwidth][l]{(a) Long-Horizon Navigation}\par\nobreak
    \includegraphics[width=\textwidth,trim=0bp 125.46bp 0bp 27.36bp,clip]{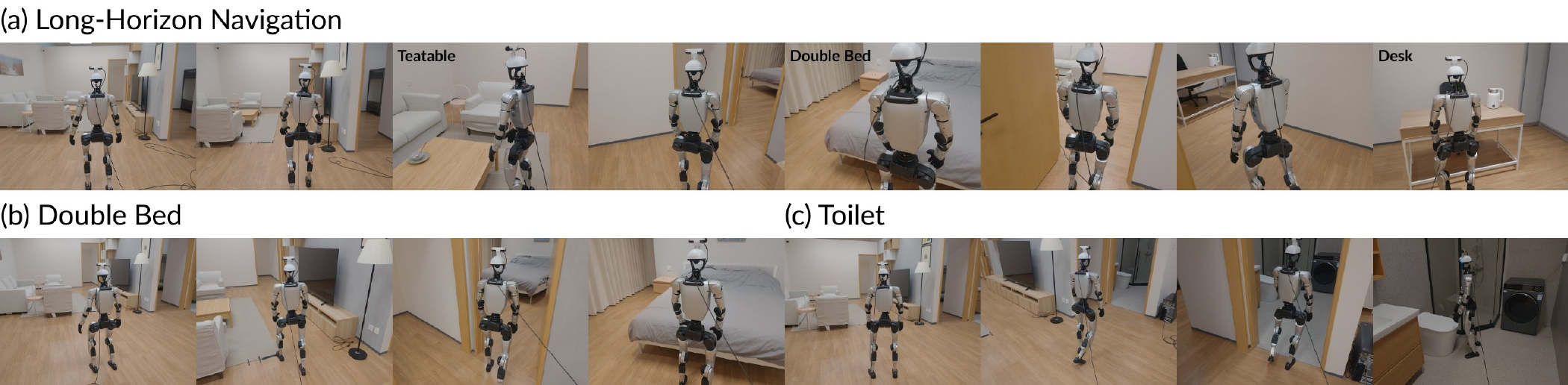}\par\nobreak
    \makebox[\textwidth][l]{\makebox[0.5\textwidth][l]{(b) Double Bed}(c) Toilet}\par\nobreak
    \includegraphics[width=\textwidth,trim=0bp 0bp 0bp 152.82bp,clip]{figures/optimized/beijing_real_navigation.pdf}\par}
    \caption{\textbf{Household navigation on Unitree G1.}
    (a) A long-horizon run visits the teatable, double bed, and desk.
    (b,c) Single-goal runs to the double bed and toilet.
    Selected frames retain intermediate walking, with goal locations
    labeled in keyframes.}
    \label{fig:real-household}
\end{figure*}

After scene-specific single-goal training with our data engine, we test
Unitree G1 from random starts (\cref{tab:m3-real-robot}).
Tests cover five household destinations, three targets in three layouts
differing from training, and one zero-shot long-horizon instruction: ``I left my
reading notes on the teatable. Please bring them to the double bed so I can
read before sleep, and once I am done, leave them on the desk for tomorrow
morning.'' Five trials per condition yield 100\% household single-goal success
and 93.3--100\% across reconfigured targets; the three-destination
reading-notes task succeeds in four of five trials.
These results demonstrate sim-to-real navigation under varied starts and
object layouts, and multi-goal execution from single-goal training.
\Cref{fig:real-household,fig:real-layouts} show representative rollouts.
Real-robot demonstrations are provided in the accompanying supplementary video.

\input{tables/m3_real_robot_results}

\begin{figure}[t]
    \centering
    \includegraphics[width=\columnwidth]{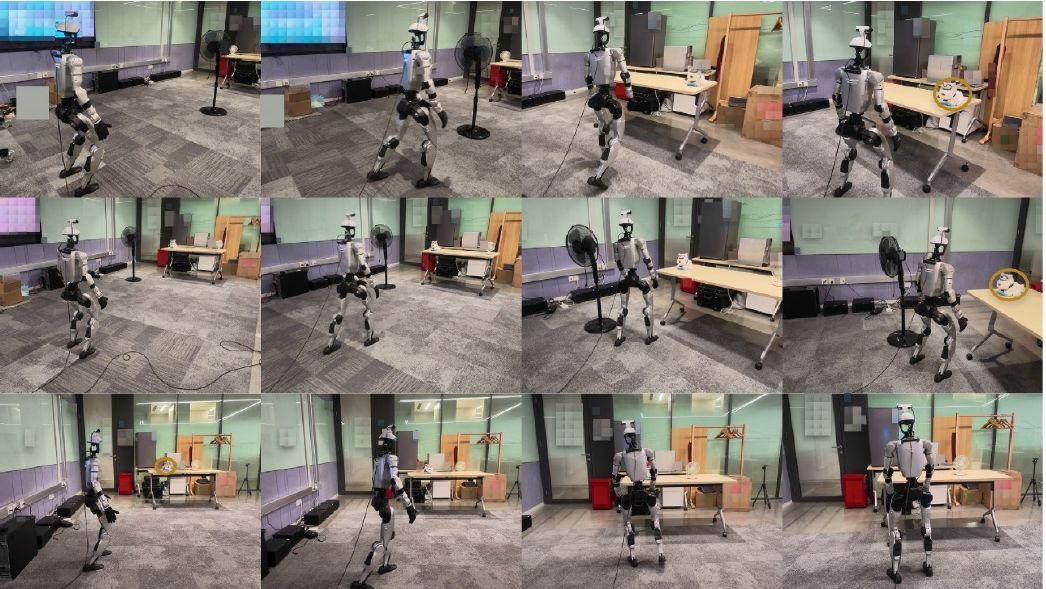}
    \caption{\textbf{Toy-dog navigation across three reconfigured layouts.}
    Each row shows the robot approaching the toy dog in one layout.
    Circles identify the target where visible; background text is masked.}
    \label{fig:real-layouts}
\end{figure}

Joint training on our simulated navigation data and real-robot manipulation
data collected through SONIC teleoperation also enables navigation to a
table followed by pushing a toy dog (\cref{fig:real-vla}), demonstrating
compatibility between the two data sources.

\begin{figure}[t]
    \centering
    \includegraphics[width=\columnwidth]{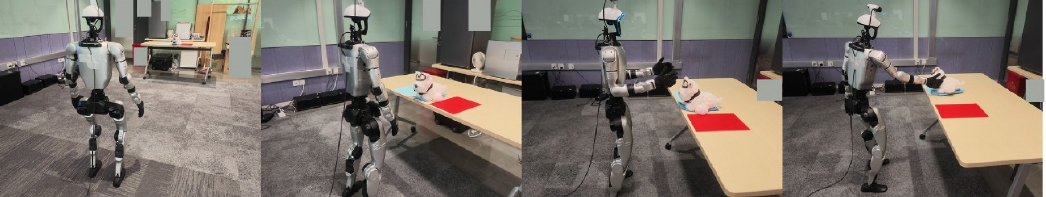}
    \caption{\textbf{Navigation followed by tabletop interaction.}
    Instruction: ``Push the toy dog from the middle region to the blue region.''
    The robot approaches the table, positions itself, and initiates a push.}
    \label{fig:real-vla}
\end{figure}

\subsection{Ablation Studies}
\label{sec:exp:ablations}

\subsubsection{Privileged Motion Tracking}
\label{sec:exp:psonic-tracking}

We evaluate SONIC and P-SONIC on 100 procedurally randomized 10~m
routes: 25 each with gentle curves, left turns, right turns, and S-curves.
Error measures final-frame horizontal distance to the ARDY reference.
P-SONIC reduces mean error by 96.4\% (\cref{tab:psonic-tracking}).

\input{tables/psonic_tracking_results}

\subsubsection{Navigation Action Representations}
\label{sec:exp:action-representations}

We compare waypoint-only, motion-token-only, and joint prediction on
522 episodes (\cref{tab:action-representations}). Under the evaluated interfaces,
the motion-token variant achieves higher success, lower joint-position jerk,
and fewer falls than the waypoint variant. Joint prediction adds waypoints
for geometric supervision, raising SR$_{0.5}$ by 4.41 percentage points and
lowering CR relative to motion-token-only prediction, with no falls.
These results support combining motion-token execution with auxiliary
geometric prediction for accurate, stable navigation.

\input{tables/action_representation_results}

\subsubsection{Geometric Supervision}
\label{sec:exp:a1}

We isolate geometric supervision by training the same joint-prediction model
with and without this term, keeping initialization, data, and training budget
fixed (\cref{tab:a1-navmesh}). Applying geometric supervision to auxiliary
waypoints raises SR$_{0.5}$ by 4.98 percentage points and lowers CR, with
similar SR$_{1.5}$ and joint-position jerk. These results show that constraining
the predicted geometry improves precise goal reaching and adherence to
traversable space during motion-token execution, without compromising
motion smoothness.

\input{tables/a1_navmesh_ablation}

\subsubsection{VLA Update Frequency}
\label{sec:exp:update-frequency}

We evaluate Cerebellum update rates of approximately 2, 5, and 8~Hz over
522 episodes, with the SONIC Brainstem fixed at 50~Hz
(\cref{tab:update-frequency}). Success rates remain similar; slower updates
reduce joint-position jerk, while approximately 8~Hz yields the lowest CR
and no falls. These results support low-rate visual action prediction coupled
with high-rate whole-body execution, with a trade-off between motion
smoothness and route adherence.

\input{tables/update_frequency_results}

\FloatBarrier
\endgroup

%% file: tables/m1_short_sim_results.tex
\begin{table}[!htb]
\centering
\footnotesize
\setlength{\tabcolsep}{2pt}
\begin{tabular}{@{}lcccccc@{}}
\toprule
Method & SR$_{0.5}$ $\uparrow$ & SR$_{1.5}$ $\uparrow$ & NE $\downarrow$ & CR $\downarrow$ & $J_p$ $\downarrow$ & Falls $\downarrow$ \\
 & (\%) & (\%) & (m) & (\%) & (m/s$^3$) & (count) \\
\midrule
OmniNav~\cite{xue2025omninav} & 36.0 & 69.0 & 1.8 & 19.5 & 109.1 & 7 \\
DualVLN~\cite{wei2025dualvln} & 18.6 & 48.7 & 2.9 & 25.8 & 128.3 & 91 \\
\method & \textbf{52.3} & \textbf{75.7} & \textbf{1.7} & \textbf{8.6} & \textbf{75.6} & \textbf{0} \\
\bottomrule
\end{tabular}
\caption{Comparison of single-goal humanoid navigation on 522 episodes
across 20 held-out scenes.}
\label{tab:m1-short-sim}
\end{table}

%% file: tables/system_results.tex
\begin{table}[!htb]
\centering
\footnotesize
\setlength{\tabcolsep}{2pt}
\begin{tabular}{@{}lccccc@{}}
\toprule
Variant & Task SR$_{1.5}$ & Progress & Time & NE & OSR \\
 & (\%) $\uparrow$ & (\%) $\uparrow$ & (s) $\downarrow$ & (m) $\downarrow$ & (\%) $\uparrow$ \\
\midrule
\multicolumn{6}{@{}l}{\textit{VLM-based system}} \\
Synchronous & 73.3 & \textbf{82.5} & 97.8 & 1.2 & 88.9 \\
w/o Task management & 0.0 & 28.0 & -- & 4.6 & 15.6 \\
w/o Arrival verification & 55.6 & 68.9 & \textbf{80.4} & 2.0 & 77.8 \\
Full system & \textbf{75.6} & 82.2 & 82.1 & 1.7 & \textbf{91.1} \\
\midrule
\multicolumn{6}{@{}l}{\textit{Codex-based system}} \\
Native & 55.6 & 76.0 & 154.5 & 1.2 & 82.2 \\
+ Harness & 64.4 & 73.3 & 141.8 & 1.9 & 82.2 \\
+ Harness + tools & 71.1 & 78.5 & 97.9 & \textbf{1.1} & \textbf{91.1} \\
\bottomrule
\end{tabular}
\caption{Comparison of long-horizon humanoid navigation on 45 tasks.}
\label{tab:system}
\end{table}

%% file: tables/m3_real_robot_results.tex
\begin{table}[!htb]
\centering
\footnotesize
\setlength{\tabcolsep}{3pt}
\begin{tabular}{@{}lcccc@{}}
\toprule
Target or task & Layouts & Trials & Successes & Success (\%) \\
\midrule
\multicolumn{5}{@{}l}{\textit{Household single-goal}} \\
Desk & 1 & 5 & 5 & 100.0 \\
Double bed & 1 & 5 & 5 & 100.0 \\
Coffee table & 1 & 5 & 5 & 100.0 \\
Toilet & 1 & 5 & 5 & 100.0 \\
Bunk bed & 1 & 5 & 5 & 100.0 \\
\midrule
\multicolumn{5}{@{}l}{\textit{Household long-horizon}} \\
Reading-notes task & 1 & 5 & 4 & 80.0 \\
\midrule
\multicolumn{5}{@{}l}{\textit{Reconfigured single-goal}} \\
Floor-standing fan & 3 & 15 & 14 & 93.3 \\
Small fan & 3 & 15 & 14 & 93.3 \\
Toy dog & 3 & 15 & 15 & 100.0 \\
\bottomrule
\end{tabular}
\caption{Real-robot navigation with Unitree G1 in household and reconfigured
environments.}
\label{tab:m3-real-robot}
\end{table}

%% file: tables/psonic_tracking_results.tex
\begin{table}[!htb]
\centering
\footnotesize
\setlength{\tabcolsep}{5pt}
\begin{tabular}{@{}lrrrr@{}}
\toprule
Tracker & Mean $\downarrow$ & Median $\downarrow$ & P95 $\downarrow$ & Max. $\downarrow$ \\
 & (cm) & (cm) & (cm) & (cm) \\
\midrule
SONIC & 48.65 & 47.59 & 73.66 & 94.02 \\
P-SONIC & \textbf{1.76} & \textbf{1.39} & \textbf{4.80} & \textbf{7.06} \\
\bottomrule
\end{tabular}
\caption{Final-frame horizontal position error.}
\label{tab:psonic-tracking}
\end{table}

%% file: tables/action_representation_results.tex
\begin{table}[!htb]
\centering
\footnotesize
\setlength{\tabcolsep}{2pt}
\begin{tabular}{@{}lcccccc@{}}
\toprule
Prediction & SR$_{0.5}$ $\uparrow$ & SR$_{1.5}$ $\uparrow$ & NE $\downarrow$ & CR $\downarrow$ & $J_p$ $\downarrow$ & Falls $\downarrow$ \\
 & (\%) & (\%) & (m) & (\%) & (m/s$^3$) & (count) \\
\midrule
Waypoints & 26.4 & 64.9 & 2.0 & 12.6 & 187.7 & 86 \\
Motion tokens & 47.9 & 73.9 & 1.9 & 12.3 & 80.9 & \textbf{0} \\
Joint prediction & \textbf{52.3} & \textbf{75.7} & \textbf{1.7} & \textbf{8.6} & \textbf{75.6} & \textbf{0} \\
\bottomrule
\end{tabular}
\caption{Navigation action representations on 522 episodes.
Both waypoint-only and joint prediction use geometric supervision.}
\label{tab:action-representations}
\end{table}

%% file: tables/a1_navmesh_ablation.tex
\begin{table}[!tb]
\centering
\footnotesize
\setlength{\tabcolsep}{2pt}
\begin{tabular}{@{}lcccccc@{}}
\toprule
\shortstack{Geometric\\supervision} & SR$_{0.5}$ $\uparrow$ & SR$_{1.5}$ $\uparrow$ & NE $\downarrow$ & CR $\downarrow$ & $J_p$ $\downarrow$ & Falls $\downarrow$ \\
 & (\%) & (\%) & (m) & (\%) & (m/s$^3$) & (count) \\
\midrule
Without & 47.3 & 75.5 & \textbf{1.6} & 14.4 & \textbf{75.3} & 3 \\
With & \textbf{52.3} & \textbf{75.7} & 1.7 & \textbf{8.6} & 75.6 & \textbf{0} \\
\bottomrule
\end{tabular}
\caption{Ablation of geometric supervision on 522 single-goal episodes.}
\label{tab:a1-navmesh}
\end{table}

%% file: tables/update_frequency_results.tex
\begin{table}[!htb]
\centering
\footnotesize
\setlength{\tabcolsep}{2pt}
\begin{tabular}{@{}lcccccc@{}}
\toprule
Nominal rate & SR$_{0.5}$ $\uparrow$ & SR$_{1.5}$ $\uparrow$ & NE $\downarrow$ & CR $\downarrow$ & $J_p$ $\downarrow$ & Falls $\downarrow$ \\
 & (\%) & (\%) & (m) & (\%) & (m/s$^3$) & (count) \\
\midrule
$\sim$2 Hz & 51.7 & 75.3 & 1.7 & 9.2 & \textbf{53.0} & 1 \\
$\sim$5 Hz & 51.9 & 73.4 & 1.8 & 10.1 & 61.3 & 1 \\
$\sim$8 Hz & \textbf{52.3} & \textbf{75.7} & \textbf{1.7} & \textbf{8.6} & 75.6 & \textbf{0} \\
\bottomrule
\end{tabular}
\caption{VLA update-rate ablation on 522 episodes.}
\label{tab:update-frequency}
\end{table}

%% file: sections/5_conclusion.tex
\section{Conclusion}
\label{sec:conclusion}

We presented \method, an asynchronous framework for long-horizon humanoid
navigation. Task-progress management and arrival verification coordinate
successive goals, while a motion-token VLA and a frozen whole-body decoder
support whole-body navigation. A physics-grounded data engine supplies
route-conditioned demonstrations with aligned visual observations and
humanoid action supervision. Simulation experiments show improved goal
reaching, smoother motion, and fewer collisions and falls, while asynchronous
coordination extends policies trained on single-goal demonstrations to
long-horizon tasks. Deployment on Unitree G1 further demonstrates household
single-goal navigation, zero-shot long-horizon navigation, and single-goal
navigation under object-layout changes.